\documentclass[conference]{IEEEtran}
\usepackage{enumitem}
\usepackage{tikz}
\usetikzlibrary{arrows.meta, positioning, decorations.pathreplacing, calc}
\usepackage{array}
\usepackage{booktabs}
\usepackage{graphicx}
\usepackage{rotating}
\usepackage{tabularx}      
\usepackage{booktabs}      
\usepackage{ragged2e}      
\usepackage{pifont}        
\newcommand{\cmark}{\ding{51}}
\newcommand{\xmark}{\ding{55}}
\newcolumntype{Y}{>{\RaggedRight\arraybackslash}X} 
\usepackage{amssymb}   
\usepackage{caption}   
\usetikzlibrary{arrows.meta, positioning}
\usepackage{cite}
\usepackage{array}
\usepackage{makecell}
\usepackage{amsmath,amssymb,amsfonts}
\usepackage{algorithmic}
\usepackage{graphicx}
\usepackage{textcomp}
\usepackage{xcolor}
\usepackage{url}
\usepackage{hyperref} 
\usepackage{soul}
\usetikzlibrary{arrows.meta, positioning, fit, calc}
\usepackage{balance}
\usepackage{amsthm}
\newtheorem{definition}{Definition}
\usepackage[table]{xcolor}

\definecolor{fc-input}{RGB}{222,235,247}
\definecolor{fc-pre}{RGB}{198,239,206}
\definecolor{fc-pos}{RGB}{237,226,246}
\definecolor{fc-attn}{RGB}{255,229,204}
\definecolor{fc-norm}{RGB}{224,224,224}
\definecolor{fc-ffn}{RGB}{255,242,204}
\definecolor{fc-pool}{RGB}{217,234,211}
\definecolor{fc-mlp}{RGB}{204,229,255}
\definecolor{fc-out}{RGB}{255,204,204}

\tikzset{
  >={Latex[length=1.6mm]},
  box/.style={
    draw, rounded corners=2pt, align=center,
    text width=0.9\columnwidth, 
    inner sep=3pt, line width=0.5pt
  },
  step/.style={
    draw, rounded corners=2pt, align=center,
    text width=0.82\columnwidth, 
    inner sep=3pt, line width=0.4pt, font=\footnotesize
  }
}
\def\BibTeX{{\rm B\kern-.05em{\sc i\kern-.025em b}\kern-.08em
    T\kern-.1667em\lower.7ex\hbox{E}\kern-.125emX}}
\begin{document}

\title{DETECT: Data-Driven Evaluation of Treatments Enabled by Classification Transformers\\
}

\author{
\begin{tabular}{ccc}
1\textsuperscript{st} Yuanheng Mao & 1\textsuperscript{st} Lillian Yang & 1\textsuperscript{st} Stephen Yang \\
\textit{Belmont High School} & \textit{Lexington High School} & \textit{Lexington High School} \\
Belmont, MA, USA & Lexington, MA, USA & Lexington, MA, USA \\
yuanhengm@gmail.com & lillianyang2017@gmail.com & stephenyang0520@gmail.com
\end{tabular}
\\[1em]
\begin{tabular}{cc}
1\textsuperscript{st} Ethan Shao & 2\textsuperscript{nd} Zihan Li \\
\textit{Lexington High School} & \textit{University of Massachusetts Boston} \\
Lexington, MA, USA & Boston, MA, USA \\
ethan.shao@gmail.com & zihan.li001@umb.edu
\end{tabular}
}

\maketitle

\begin{abstract}
Chronic pain is a global health challenge affecting millions of individuals, making it essential for physicians to have reliable and objective methods to measure the functional impact of clinical treatments. Traditionally used methods, like the numeric rating scale, while personalized and easy to use, are subjective due to their self-reported nature. Thus, this paper proposes DETECT (Data-Driven Evaluation of Treatments Enabled by Classification Transformers), a data-driven framework that assesses treatment success by comparing patient activities of daily life before and after treatment. We use DETECT on public benchmark datasets and simulated patient data from smartphone sensors. Our results demonstrate that DETECT is objective yet lightweight, making it a significant and novel contribution to clinical decision-making. By using DETECT, independently or together with other self-reported metrics, physicians can improve their understanding of their treatment impacts, ultimately leading to more personalized and responsive patient care.
\end{abstract}


\section{Introduction}

Chronic pain is one of the most common reasons adults seek medical care. In 2021, it affected more than 20\% of US adults, causing significantly reduced productivity\cite{ChronicPainAmongAdults}. Currently, chronic pain is commonly evaluated using self-reported metrics such as the numeric rating scale (NRS), where patients rate their pain from 0 to 10\cite{PainRatingScale}. These subjective methods are vulnerable to bias and variability, making accurate treatment assessment difficult\cite{Marty2009BackPain}. To address these challenges, we present DETECT (Data-Driven Evaluation of Treatments Enabled by Classification Transformers), a transformer-based evaluation framework that assesses chronic pain treatment success by detecting changes in activities of daily life (ADL) patterns before and after clinical intervention. 
 
This study uses sets of public benchmark and simulated patient ADL data. Real patient data collection is ongoing with a chronic pain physician at Massachusetts General Hospital (MGH), who has expressed interest in our system. We used commercial smartphones for data collection as they provide a widely accessible and patient-friendly alternative to clinical wearable sensors. We created a mobile application for iOS and Android, ActiPain Tracker, to allow patients to collect data in both clinical and home environments. The app demonstrated feasibility and ease of use in real-world clinical settings when tested with patient data collection at MGH. The physician expressed further interest in expanding the app's usage once it is released for free on the App Store. This low-cost and scalable approach is valuable for treatment assessment and general functional health monitoring.

\begin{figure}[t]
    \centering
    \includegraphics[width=0.85\linewidth]{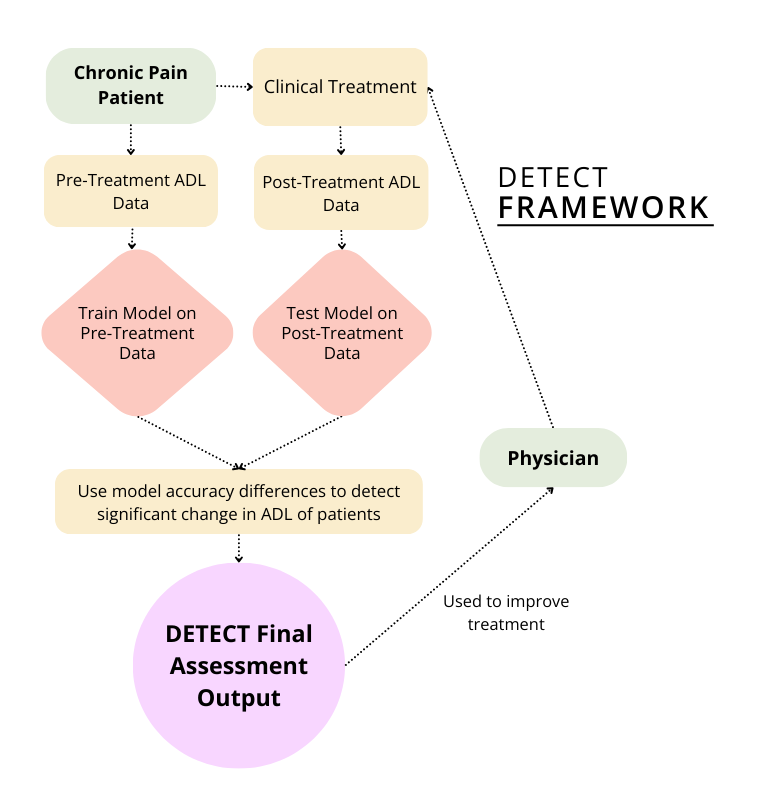} 
    \caption{The DETECT Framework compares pre- and post-treatment activity data using a transformer model to evaluate treatment impact, guiding physicians to improve patient care.}
    \label{Fig1}
\end{figure}

\begin{figure*}[t]
    \centering
    \includegraphics[width=0.85\textwidth]{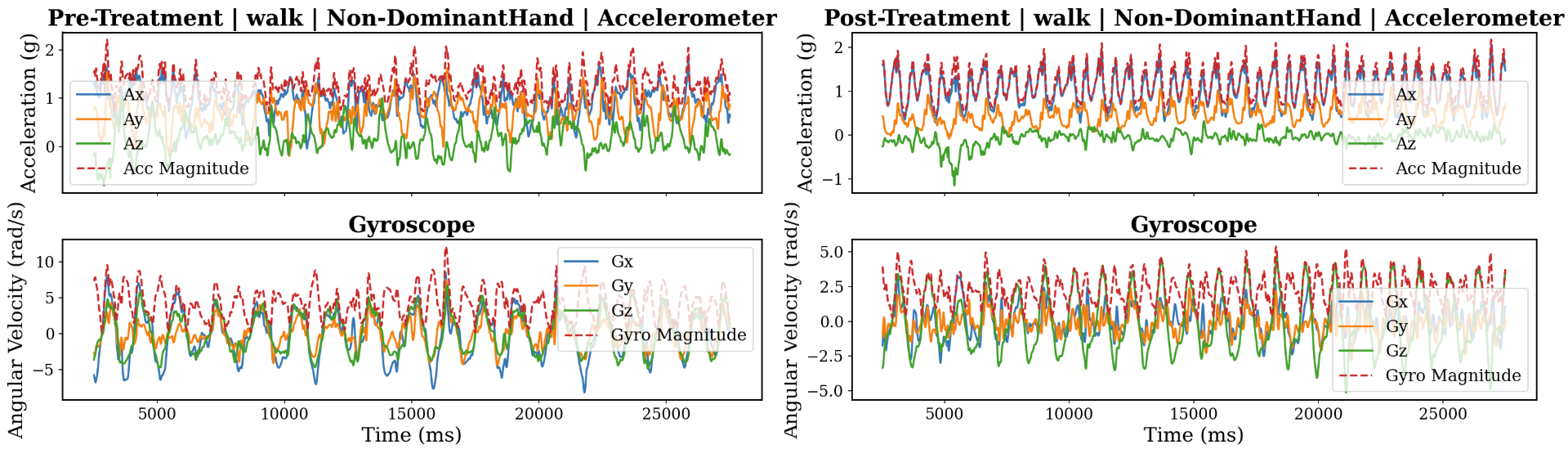}
    \caption{Graphs displaying accelerometer and gyroscope measurements from a sample patient's nondominant hand during walking, collected before and after successful treatment.}
    \label{Fig2}
\end{figure*}

Our methodology, illustrated in Fig.~\ref{Fig1}, is rooted in a transformer-based classification model trained on pre-treatment data and tested on post-treatment data. For change detection, if post-treatment data differ significantly, the model’s classification accuracy will decrease, signaling behavioral shifts potentially due to treatment. This is justified since, as shown in Fig.~\ref{Fig2}, the movement patterns of patients before and after treatment are visibly different, to the extent that a transformer model can decisively detect such a shift. In Fig.~\ref{Fig2}, pre-treatment data are shown on the left and post-treatment data on the right. Upon comparison, post-treatment data exhibit stabler acceleration and angular velocity with reduced variability, indicating improved gait regularity and motor control. This framework provides physicians with an objective measure of functional improvement, supplementing or replacing subjective scales and allowing physicians to provide more personalized treatment. 

Thus, our main novel contributions are as such:
\begin{enumerate}[itemsep=5pt,topsep=5pt]
    \item Personalization: personalized treatment evaluation by adapting analysis to each patient's unique ADL patterns.
    \item Accessibility: increasing accessibility of treatment assessment through a free mobile app allowing for home-based monitoring without specialized clinical sensors.
    \item Objectivity: creating a data-driven approach for assessing treatment impact that offers greater objectivity than traditional self-reported pain scales.
\end{enumerate}

\section{Related Work}
Many methods are used to evaluate differences in pain intensity and mobility after clinical treatment. With recent technological advancements, researchers are increasingly applying machine learning (ML) and artificial intelligence (AI) for healthcare and pain diagnostics\cite{Lotsch2022AIMLScientometric}. Advances in mobile technology have also enabled researchers to use smartphone and wearable sensor data to monitor chronic pain and associated behavioral impairments. In particular, accelerometers are widely used to capture movement intensity and gait characteristics relevant to pain-related limitations\cite{Xing2024AIPainMedicine}. 

The most traditionally used method for pain assessment is the 11-point NRS, where patients rate their pain from 0 (“no pain”) to 10 (“worst imaginable pain”). Another common method is through questionnaires such as the Roland-Morris Disability Questionnaire (RMDQ) and the Patient-Reported Outcomes Measurement Information System (PROMIS), which measures daily functional limitations\cite{roland1983rmdq, cella2007promis}. While these methods are easily accessible, they rely solely on patient memory and perception, making them subjective and prone to bias. 

In addition to patient self-reporting, there are also more objective methods such as autonomic nervous system (ANS) measurements, which use physiological assessments (e.g., heart rate variability, skin conductance) that reflect autonomic activity potentially linked to pain\cite{zygmunt2010}. These can provide objective signals but require specialized sensors, making them resource-intensive. Additionally, neuroimaging-based biomarkers use brain imaging techniques such as fMRI or PET scans to identify neural activity patterns correlated with pain\cite{vandermiesen2019neuroimaging}. While objective, these are also very resource-intensive.

All of these methods provide insights into treatment assessment, but are limited in several important ways. Therefore, we propose DETECT: an objective, accessible, and easily deployable system which, as shown in Table~\ref{tab:compare}, covers more vital areas than other methods. By using smartphone-based motion data collected during ADL, DETECT can reveal treatment-related behavioral changes without requiring expensive equipment, extensive clinical resources, or subjective self-reporting. 

\newcommand{\rot}[1]{\rotatebox[origin=c]{90}{#1}}
\begin{table}[htbp]
\caption{Comparison across DETECT and various methods.}
\label{tab:compare}
\begin{center}
\renewcommand{\arraystretch}{1.35}
\setlength{\tabcolsep}{8pt}
\begin{tabular}{
    l !{\vrule width 1pt} c c c c c || c
}
\toprule
& \rot{NRS} & \rot{RMDQ} & \rot{PROMIS} & \rot{ANS} & \rot{Biomarkers} & \rot{DETECT} \\
\midrule
Personalization & \checkmark & \checkmark & \checkmark & \checkmark & \checkmark & \checkmark \\
Accessibility   & \checkmark & \checkmark & \checkmark &            &            & \checkmark \\
Objectivity     &            &            &            & \checkmark & \checkmark & \checkmark \\
\bottomrule
\end{tabular}
\end{center}
\end{table}

\section{Methodology}

\subsection{Accessibility}    
\begin{figure}[ht]
    \centering
    \begin{minipage}[b]{0.45\columnwidth}
        \centering
        \includegraphics[width=0.9\linewidth]{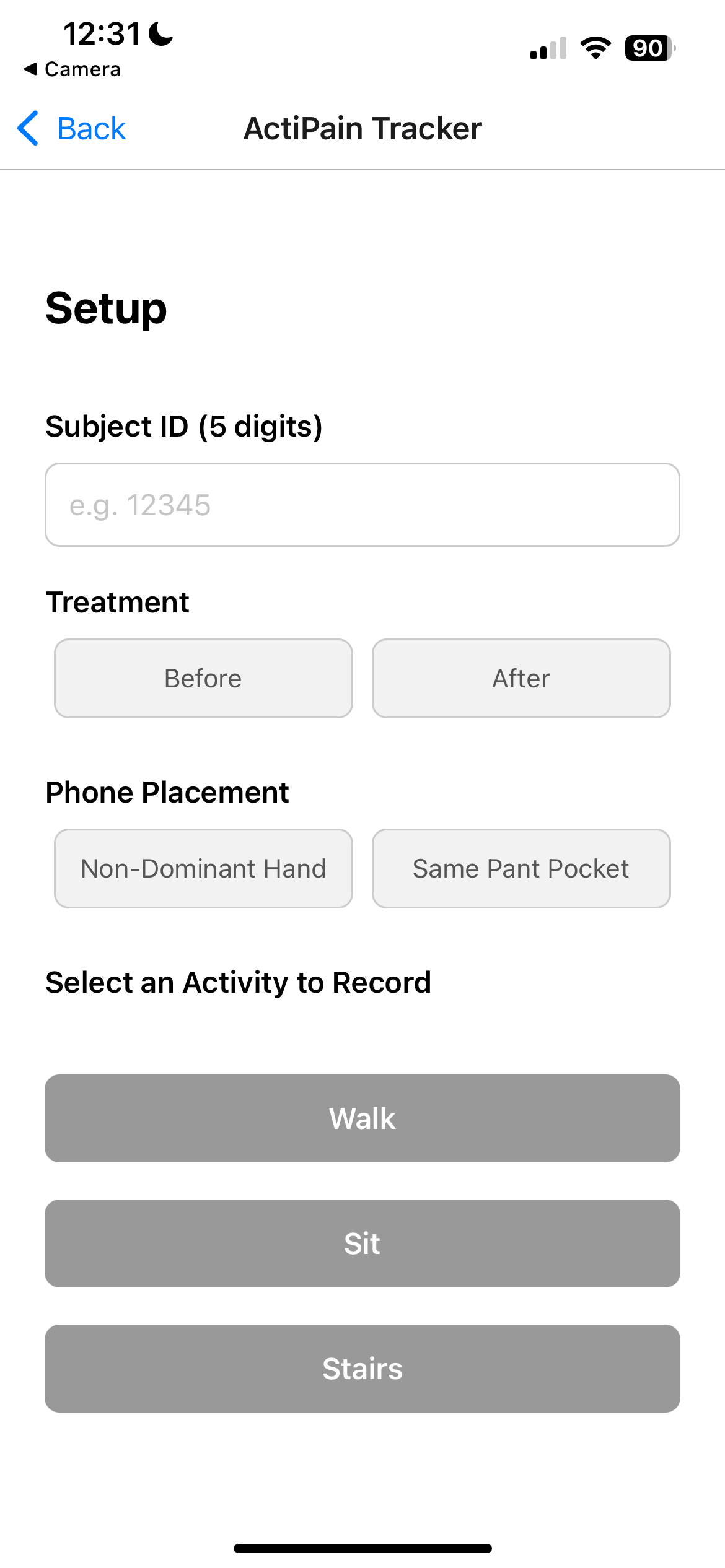}
    \end{minipage}
    \hfill
    \begin{minipage}[b]{0.45\columnwidth}
        \centering
        \includegraphics[width=0.9\linewidth]{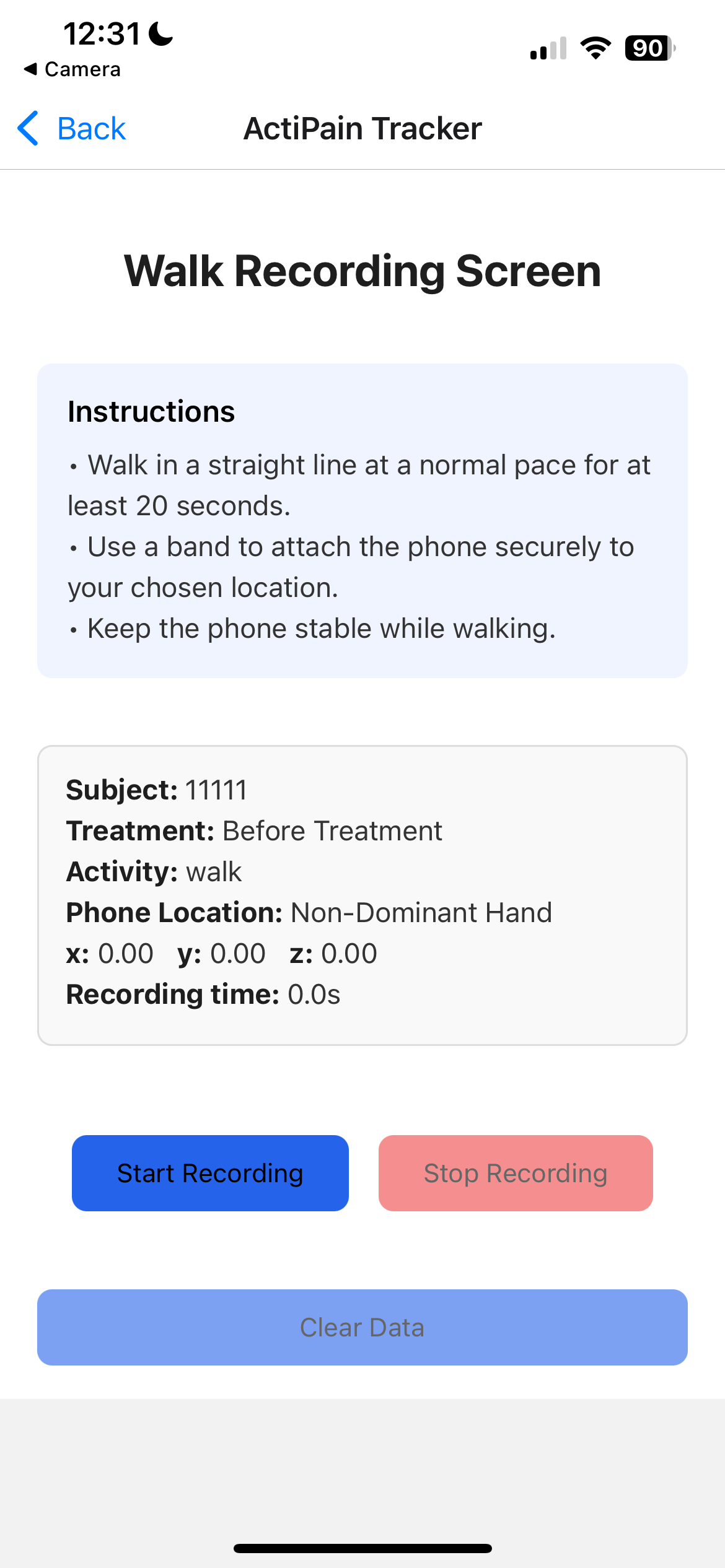}
    \end{minipage}
    \caption{Mobile app user interface. Patient information screen (left) allows entry of subject ID, treatment phase, phone placement, and activity selection. "Walk" activity recording screen (right) displays instructions and controls for data collection.}
    \label{Fig3}
\end{figure}

    We developed a mobile application, ActiPain Tracker, using React Native with Expo for compatibility on both Android and iOS. The app, displayed in Fig.~\ref{Fig3}, is designed to be freely available on app stores and open to the public, allowing for increased accessibility and easy data collection. Data were collected using embedded accelerometers and gyroscopes. Accelerometers measure linear acceleration while gyroscopes capture angular velocity. Together, they allow us to track subtle changes in gait or posture that could be associated with chronic pain. These factors make DETECT highly accessible, supporting patients beyond restrictive clinical settings.  

\subsection{The DETECT Framework}
    The DETECT Framework is built on a transformer-based deep learning model for activity classification. The model is trained and validated on ADL pre-treatment data and tested on post-treatment data. To determine treatment impact, the model’s classification accuracy on each patient's pre-treatment data is compared to its performance on their post-treatment data. If post-treatment movement remains similar to pre-treatment movement, the model’s classification accuracy will remain high. However, if movement patterns change significantly due to treatment effects, the model will struggle to classify the new data accurately, leading to a measurable drop in accuracy.

    This framework represents a novel way to assess treatment outcomes objectively. It complements traditional self-reported measures such as the NRS by providing a data-driven and unbiased assessment of treatment efficacy. By combining changes in model accuracy with patient-reported outcomes, the DETECT Framework offers a more robust and accessible measure of treatment impact.

To formalize the DETECT Framework, let $X = \{x_1,\dots,x_n\}$ be a sequence of sensor samples, where each $x_i \in \mathbb{R}^{d}$ represents a $d$-dimensional feature vector at time $i$. Each sample has a corresponding activity label
\begin{equation}
y_i \in \mathcal{C} , \label{eq:labels}
\end{equation}
where $\mathcal{C}= \{c_1, c_2, \dots, c_K\}$ is the set of $K$ activity classes.  

\begin{figure}[t]
    \centering
    \includegraphics[width=0.85\linewidth]{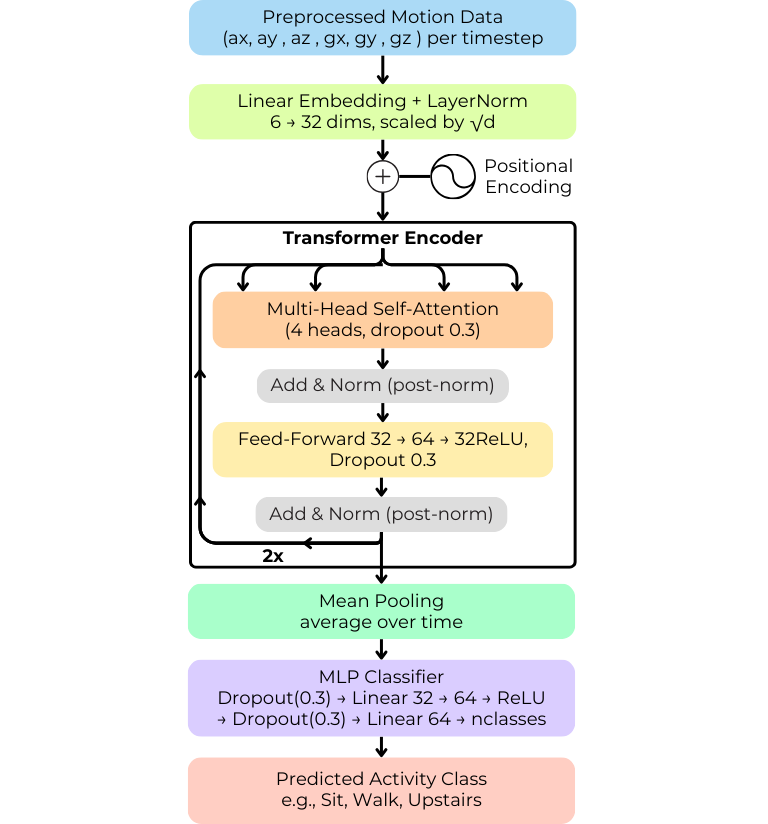} 

\caption{Design and architecture of the transformer-based activity classification model used in the DETECT Framework.}
\label{Fig4}
\end{figure}

We use a transformer model, outlined in Fig.~\ref{Fig4}, for activity classification. Each input vector is linearly projected into an $m$-dimensional latent space and augmented with positional encoding. The resulting sequence is processed by a standard transformer encoder with $L$ stacked layers \cite{vaswani2017attention}, leading to contextualized representations $H^{(L)} = \{h_1^{(L)},\dots,h_n^{(L)}\}$. A global average pooling step produces $\bar{h} = n^{-1}\sum_i h_i^{(L)}$, which is mapped to class probabilities through a softmax classifier. Model parameters $\Theta$ are optimized with cross-entropy loss with label smoothing ( $\epsilon = 0.1$):
\begin{equation}
\mathcal{L}(\Theta) = -\,\frac{1}{N} \sum_{j=1}^{N} \sum_{k=1}^{K} 
\Big[(1-\epsilon)\,\mathbf{1}\{y_j=c_k\} + \frac{\epsilon}{K}\Big] \, \log \hat{y}_{j,k}.
\label{eq:loss_smoothed}
\end{equation}
Optimization is performed using AdamW with weight decay, and the learning rate is scheduled with a warm-up phase followed by cosine decay.  For reproducibility, all experiments were conducted with fixed random seeds (SEED = 42) across Numpy, Python, and Pytorch.

We calculate the Treatment Effect Score (TES) for each patient $p$ based on Definition~\ref{def:tes}, and use the TES to detect the impact of treatment given by a doctor. A larger $\mathrm{TES}_{p}$ indicates a greater behavioral shift after treatment.

\begin{definition}[Treatment Effect Score]\label{def:tes}
Given a model $M$ trained solely on pre-treatment data, for a patient $p$, the difference between the classification accuracy of the patient $p$’s pre-treatment ADL data ($\operatorname{Acc}_{p,\mathrm{pre}}$) and that of the post-treatment ADL data ($\operatorname{Acc}_{p,\mathrm{post}}$) is defined as the Treatment Effect Score (TES), where
\begin{equation}
\mathrm{TES}_{p} = \operatorname{Acc}_{p,\mathrm{pre}} - \operatorname{Acc}_{p,\mathrm{post}}. \label{eq:tes}
\end{equation}
\end{definition}

To determine whether this shift shows a functional and clinically meaningful treatment effect, we compute the mean TES across patients who showed improvement in their NRS scores. The mean TES defines the pilot threshold:
\begin{equation}
\mathrm{TES}_{\mathrm{threshold}} = \frac{1}{|P_{\mathrm{NRS+}}|} \sum_{p \in P_{\mathrm{NRS+}}} \mathrm{TES}_{p}, \label{eq:tes-threshold}
\end{equation}
where $P_{\mathrm{NRS+}}$ is the set of patients with meaningful NRS improvement, defined as an NRS reduction of at least $33\%$ or $\geq 2$ points\cite{salaffi2004minimal}. Patients with $\mathrm{TES}_{p} \geq \mathrm{TES}_{\mathrm{threshold}}$ are considered to display significant treatment effects.

This approach establishes the DETECT Framework using patient-reported outcomes rather than only relying on an arbitrary cutoff. The resulting $\mathrm{TES}_{\mathrm{threshold}}$ can serve as an initial pilot threshold for assessing treatment impact for future patients. By using the TES of DETECT together with or in place of the NRS, physicians can gain a more objective and comprehensive evaluation of treatment outcomes.
\section{Experimental Results}

\subsection{Data Collection}
Real patient data collection is ongoing with a chronic pain doctor at MGH because Institutional Review Board (IRB) approval is still pending. As such, we designed a simulated dataset to mimic patient activity before and after treatment. Each volunteer was assigned NRS pain values for pre- and post-treatment conditions, which were then paired with ADLs such as walking, sitting, and climbing stairs. Higher-intensity activities (e.g., running or jumping) were excluded as they are often infeasible for chronic pain patients. Up and downstairs were collected together due to highly similar data patterns. 

To strengthen our dataset, each activity, pre- and post-treatment, was recorded for two 30~s trials each in two locations of the device: the nondominant hand and the same pant pocket. Prior work has shown that sensor placement influences recognition accuracy, with nondominant hand placement providing more consistent results than dominant hand\cite{Gjoreski2016}. Including the same pant pocket location allowed us to capture a stable and realistic scenario for patients who may not be able to hold their phones while performing daily activities. 

In addition to simulated data, we also used public benchmark data. For public data, we used the activities sit, upstairs, and walk from KU-HAR: An Open Dataset for Human Activity Recognition\cite{NahidKUHAR2021} and from the IMU-based Human Activity Recognition Dataset\cite{TahirIMU2025}. 

\subsection{Data Processing}
    Data were sampled at 100~Hz for the simulated and KU-HAR data and at 50~Hz for the IMU data. To reduce noise and remove transitional movements, we trimmed the first and last 2.5~s of our data before segmentation. Each recording was then segmented into 1~s windows of 100 samples each. We applied a sliding window with 50\% overlap, meaning a step size of 50 samples. We extracted six features, which were the x, y, and z components of linear acceleration and angular velocity (i.e. $a_x, a_y, a_z, g_x, g_y, g_z$). The features were z-score normalized using the training set and applied to the data.

\subsection{Model Training}
    With the DETECT Framework, we trained solely on pre-treatment data to establish a personalized baseline for the patients. The data were split into 80\% training and 20\% validation sets using stratified sampling. The IMU dataset was also smaller than the KU-HAR, so it was run using 5-fold cross-validation.
    
    Training ran for 100 epochs using the AdamW optimizer with initial learning rate $=0.001$, weight decay $=10^{-4}$, and batch size of 32. The loss function was cross-entropy with label smoothing of $\epsilon=0.1$. We applied gradient clipping with $\ell_2$ norm capped at 1.0, and a learning rate scheduler, consisting of linear warmup followed by cosine decay. Dropout layers were included in both the encoder and classifier to improve generalization.
    
    An implementation of our framework has been permanently archived on Zenodo and is available at \href{https://doi.org/10.5281/zenodo.17096042}{doi:10.5281/zenodo.17096042}.

\subsection{Simulated Pain Data Results}    
    We observed significant differences between train and test accuracies using our simulated patient data. We trained the activity classification model on all the pre-treatment data together. On pre-treatment data, the model achieved 98.03\% mean accuracy. However, when testing on each simulated patient's post-treatment data, the test accuracy was lower than the training accuracy by varying amounts. These results, displayed in Table~\ref{tab:results}, illustrate how four of the TES values, 11.28, 15.80, 12.76, and 14.63, exceeded the $\mathrm{TES}_{threshold}$ of 11.10, showing evidence of significant treatment impact.
    
    Additionally, as also shown in Table~\ref{tab:results}, DETECT and the NRS had the same output on seven, or 87.50\%, of the eight trials. The consistency of these assessments emphasizes the strength of the DETECT Framework in successfully determining treatment impact. 

\begin{table*}[!t]
\caption{Per-patient DETECT and NRS results.}
\label{tab:results}
\centering
\scriptsize
\setlength{\tabcolsep}{5pt}
\renewcommand{\arraystretch}{1.05}

\resizebox{\textwidth}{!}{%
\begin{tabular}{|c|c|c|c|c|c|c|c|c|}
\hline
\textbf{Patient ID} &
\shortstack{\textbf{Pre-}\\\textbf{treatment accuracy (\%)}} &
\shortstack{\textbf{Post-}\\\textbf{treatment accuracy (\%)}} &
$\mathbf{TES}_{\text{patient}}$ &
$\mathbf{TES}_{\text{threshold}}$ &
\shortstack{\textbf{Pre-}\\\textbf{treatment NRS}} &
\shortstack{\textbf{Post-}\\\textbf{treatment NRS}} &
\shortstack{\textbf{Sig.}\\\textbf{NRS}} &
\shortstack{\textbf{Sig.}\\\textbf{DETECT}} \\
\hline
12345 & 98.61 & 87.33 & 11.28 & 11.10 & 5 & 1 & \underline{\cmark} & \underline{\cmark} \\
21000 & 98.61 & 82.81 & 15.80 & 11.10 & 6 & 2 & \underline{\cmark} & \underline{\cmark} \\
31000 & 97.05 & 92.88 & 4.17 & 11.10 & 5 & 5 & \underline{\xmark} & \underline{\xmark}  \\
41000 & 98.98 & 86.22 & 12.76 & 11.10 & 7 & 4 & \underline{\cmark} & \underline{\cmark} \\
51000 & 97.80 & 89.12 & 8.69 & 11.10 & 3 & 2 & \underline{\xmark} & \underline{\xmark} \\
61000 & 96.94 & 82.31 & 14.63 & 11.10 & 5 & 3 & \underline{\cmark} & \underline{\cmark} \\
71000 & 97.96 & 96.94 & 1.02 & 11.10 & 2 & 0 & {\cmark} & {\xmark}  \\
91000 & 98.26 & 93.75 & 4.51 & 11.10 & 4 & 3 & \underline{\xmark} & \underline{\xmark}  \\
\hline
\end{tabular}
}

\vspace{0.5em}

\resizebox{\textwidth}{!}{%
\begin{tabular}{|l|r|r|r|r|r|r|}
\hline
\textbf{Summary Across Patients (n = 8)} &
\textbf{Pre Acc. (\%)} &
\textbf{Post Acc. (\%)} &
\textbf{TES (Patient)} &
\textbf{Pre NRS} &
\textbf{Post NRS} &
\textbf{Consistency Rate (\%)} \\
\hline
\textbf{Mean (SD)} &
98.03 (0.74) &
88.92 (5.27) &
9.11 (5.40) &
4.63 (1.60) &
2.50 (1.60) &
87.50 \\
\hline
\textbf{95\% CI} &
[97.52, 98.54] &
[85.27, 92.57] &
[5.37, 12.85] &
[3.52, 5.74] &
[1.39, 3.61] &
-- \\
\hline
\end{tabular}%
}

\vspace{0.6em}
\begin{flushleft}
\footnotesize
{\cmark} = significant improvement, {\xmark} = no improvement. The lower panel summarizes mean (SD) and 95\% CI across patients. Underlined marks indicate that NRS and DETECT results were consistent.
\end{flushleft}

\end{table*}

\subsection{Public Benchmark Data Results}
    The KU-HAR and IMU datasets are both public benchmark datasets containing accelerometer and gyroscope ADL data. When training and testing the model on the KU-HAR dataset, the model achieved 99.69\% train accuracy and 98.74\% test accuracy. When doing the same for the IMU dataset, the model achieved 98.69\% mean train accuracy and 95.06\% mean test accuracy across the five folds.
    
    These high accuracies demonstrate the model's robustness in consistently identifying activities with normal data. This demonstrates that substantial drops in model accuracy during testing on post-treatment pain data are not due to model flaws. Rather, they are due to meaningful data shifts that may result from treatment impact. Therefore, instances where a patient's TES value exceeds the threshold can be interpreted as evidence of a significant treatment-related behavioral change.

\section{Discussion}

\subsection{Limitations}
    While the DETECT Framework is robust, this study has a few limitations. Although our simulated data were carefully designed to approximate real-world patient data, our findings still have to be interpreted as approximations rather than direct clinical outcomes. Second, the short duration of our study prevented analysis of long-term treatment impacts. Third, the current study focuses on sitting, walking, and stair climbing, which may not be sufficient to represent the broader range of movements experienced by individuals with chronic pain. 

\subsection{Future Works}
   Future research could expand this study by integrating the DETECT Framework directly into our app for real-time treatment assessment for patients and physicians. Second, using more diverse real-world patient data will improve generalizability. Third, broadening activity types and collecting longer-term data will further enhance the framework's effectiveness. 

\section{Conclusion}
This study demonstrates the effectiveness of the DETECT Framework for objectively evaluating chronic pain treatment impact using accelerometer and gyroscope ADL data before and after treatment. The decline in model accuracy from post-treatment data, despite a strong model, reveals that the framework can detect clear behavioral differences corresponding with patient impact due to treatment. Through this highly personalized, accessible, and objective method of treatment evaluation, DETECT provides a scalable foundation for improved chronic pain treatment evaluation strategies.
\section*{Acknowledgments}
We thank Dr. Shiqian Shen, MD, of Massachusetts General Hospital, for collecting patient data and supporting us through this project. We are grateful to Prof. Ping Chen of UMass Boston, and Prof. Wei Ding, Executive Director of the Paul English Applied AI Institute, for their mentorship. We thank the Youth STEAM Initiative, a nonprofit high school organization that supported us throughout this project.


\bibliographystyle{IEEEtran}
\bibliography{references}
\end{document}